\pdfoutput=1

\documentclass{article}

\usepackage{ijcai24}

\usepackage{booktabs}
\usepackage{times}
\usepackage{soul}
\usepackage{url}
\usepackage[hidelinks]{hyperref}
\usepackage[utf8]{inputenc}
\usepackage[small]{caption}
\usepackage{graphicx}
\usepackage{amsmath}
\usepackage{amsthm}

\usepackage{algorithm}
\usepackage{algorithmic}
\usepackage[switch]{lineno}

\usepackage{multicol}
\usepackage{multirow}
\usepackage{microtype}
\usepackage{latexsym}
\DeclareMathOperator*{\argmax}{arg\,max}

\usepackage{amsfonts,amssymb}
\usepackage{svg}

\title{Domain-Hierarchy Adaptation via Chain of Iterative Reasoning for Few-shot Hierarchical Text Classification}

\author{
    Ke Ji$^1$
    \and
    Peng Wang$^{1,2}$\thanks{Corresponding author} \and
    Wenjun Ke$^{1,2}$\and
    Guozheng Li$^1$\and \\
    Jiajun Liu$^1$\and 
    Jingsheng Gao$^3$\and
    Ziyu Shang$^1$
    \\
    \affiliations
    $^1$School of Computer Science and Engineering, Southeast University\\
    $^2$Key Laboratory of New Generation Artificial Intelligence Technology and Its \\
    Interdisciplinary Applications (Southeast University), Ministry of Education \\
    $^3$School of Electronic Information and Electrical Engineering, Shanghai Jiao Tong University
    \emails
    \{keji, pwang, kewenjun, gzli, jiajliu, ziyus1999\}@seu.edu.cn,
    gaojingsheng@sjtu.edu.cn
}

\begin{document}

\maketitle

\begin{abstract}
Recently, various pre-trained language models (PLMs) have been proposed to prove their impressive performances on a wide range of few-shot tasks.
However, limited by the unstructured prior knowledge in PLMs, it is difficult to maintain consistent performance on complex structured scenarios, such as hierarchical text classification (HTC), especially when the downstream data is extremely scarce. 
The main challenge is how to transfer the unstructured semantic space in PLMs to the downstream domain hierarchy.
Unlike previous work on HTC which directly performs multi-label classification or uses graph neural network (GNN) to inject label hierarchy, in this work, we study the HTC problem under a few-shot setting to adapt knowledge in PLMs from an unstructured manner to the downstream hierarchy.
Technically, we design a simple yet effective method named Hierarchical Iterative Conditional Random Field (HierICRF) to search the most domain-challenging directions and exquisitely crafts domain-hierarchy adaptation as a hierarchical iterative language modeling problem, and then it encourages the model to make hierarchical consistency self-correction during the inference, thereby achieving knowledge transfer with hierarchical consistency preservation.
We perform HierICRF on various architectures, and extensive experiments on two popular HTC datasets demonstrate that prompt with HierICRF significantly boosts the few-shot HTC performance with an average Micro-F1 by 28.80\% to 1.50\% and Macro-F1 by 36.29\% to 1.5\% over the previous state-of-the-art (SOTA) baselines under few-shot settings, while remaining SOTA hierarchical consistency performance.

\end{abstract}

\begin{figure}[!t]
\centering
\includegraphics[width=0.49\textwidth]{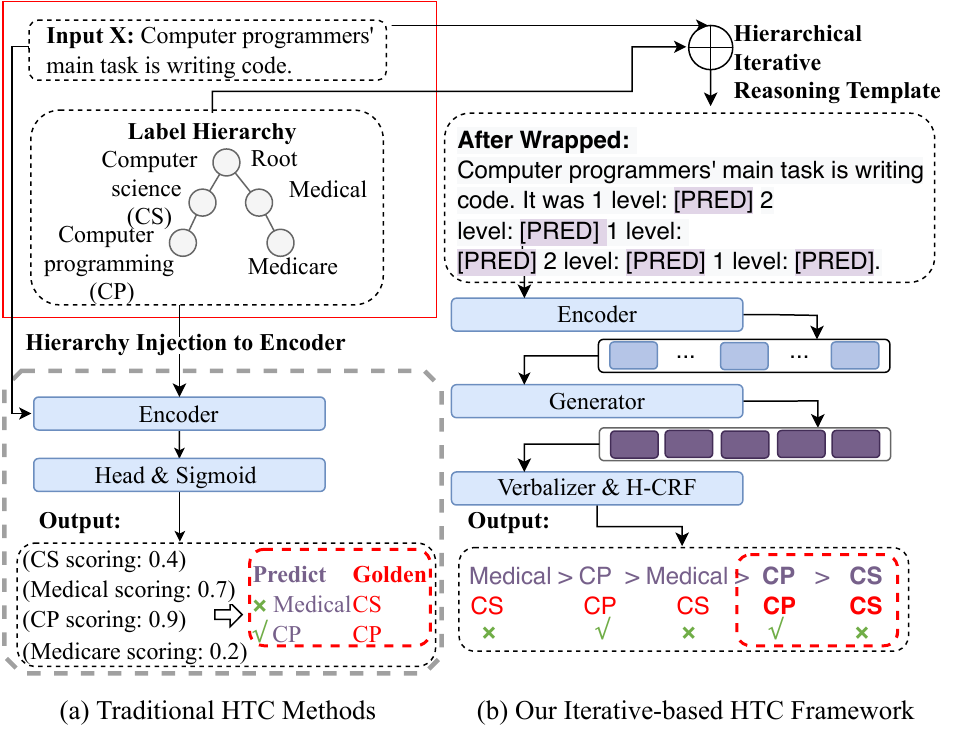}
\caption{Illustration of methods for HTC. 
The red sequence represents the golden label, and the purple sequence represents the predicted sequence.
Hierarchical inconsistency happens when the relationship between the outputs of different layers conflicts with the hierarchical dependency tree, for example, the model predicts the CP which is not the child node of its other output like Medicare.
}
\label{fig: HierICRF_Introduction}
\end{figure}

\section{Introduction}
Pre-trained Language Models (PLMs)~\cite{radford2018improving,devlin-etal-2019-bert,raffel2020exploring} have gained significant prominence for their exceptional performance across various language-related tasks, including text classification~\cite{kowsari2019text}, and relation extraction~\cite{wangfmLRE2023,li2022fastre}.

The success of PLMs mainly benefits from large-scale pre-training and sufficient downstream labeled data. 
However, when downstream labeled data is scarce, performance is greatly compromised, and it is further aggravated when we try to transfer unstructured prior knowledge in PLMs to downstream structured tasks like hierarchical text classification (HTC)~\cite{mao-etal-2019-hierarchical}.
Due to HTC's broad range of practical applications~\cite{mao-etal-2019-hierarchical}, including product categorization~\cite{cevahir2016large}, fine-grained entity typing~\cite{xu2018neural} and news classification~\cite{irsan2019hierarchical}, HTC has remained a significant research challenge over time.
Despite existing HTC methods, its complex label hierarchy and the need for extensive annotation still hinder performance in practice. 
Addressing HTC in few-shot scenarios remains an open area of research~\cite{ji2023hierarchical}.

Currently, the state-of-the-art in HTC~\cite{wang-etal-2022-incorporating,wang2022hpt,ji2023hierarchical} involves incorporating label hierarchy features into the input or output layer of a text encoder, using graph encoders or hierarchical verbalizer, as is illustrated in Figure~\ref{fig: HierICRF_Introduction}(a). 
It’s disappointing that they ignore the adaptation of unstructured prior knowledge to downstream domain-hierarchy structure and still directly consider HTC as a multi-label classification problem based on an encoder with label-hierarchy dependencies injection.
%
Inspired by the \textit{in-context learning} approach proposed by GPT-3 \cite{NEURIPS2020_1457c0d6} and prompt-based methods \cite{petroni-etal-2019-language,gao-etal-2021-making,schick-schutze-2021-exploiting,li-etal-2023-revisiting-large,li2024unlocking} that aim to bridge the gap between pre-training and downstream tasks by utilizing few hard or soft prompts to stimulate the PLMs' knowledge, \cite{wang2022hpt,ji2023hierarchical} are proposed to provide a more systematic study under low resource or few-shot settings using prompt-based methods.
However, instead of paying more attention to the inherent difference between downstream hierarchy and unstructured objective in PLMs, all these methods are just built from the perspective of how to get a hierarchical label dependency representation, leading to unstable hierarchical consistency performance.
Similar to mathematical reasoning tasks~\cite{zhu-etal-2023-solving} for which the answers are often implicit, making it difficult to deal with directly through question-answer pairs, thus the way to think about the HTC task is really unfriendly and even antithetical to the language model’s capabilities.
Thus the primary difference between these works on HTC tasks is the way they inject the label hierarchy constraint.

Despite their success, they still suffer from two limitations. 
On the one hand, previous methods are not well designed to focus on how to decompose and simplify hierarchy-based problems, on the contrary, they think of HTC in an even more complex way, and thus they are difficult to handle the hierarchical inconsistency problem.
On the other hand, considering most of the currently popular large language models are based on encoder-decoder~\cite{raffel2020exploring,chung2022scaling} or decoder-only~\cite{NEURIPS2020_1457c0d6,scao2022bloom} architectures, previous works are mostly applicable to the encoder-only architecture, which leads them to mine rich prior knowledge at a limited model scale in practical applications. 
Therefore, few studies have investigated how to efficiently handle the few-shot domain-hierarchy adaptation problem.
And few studies have tried to develop a simple and unified framework that can be flexibly deployed in any architectural model for better practical application performance.

In this work, we design a unified framework named HierICRF from the perspective of path routing that can be deployed on any transformer-based architecture to fully elicit the potential of unstructured prior knowledge in PLMs to complete downstream hierarchy tasks.
Unlike previous works that mainly focused on how to align their carefully crafted representators that incorporate label-hierarchy dependencies with the sentence semantic space, we use a language modeling routing paradigm based on hierarchical iteration to unify the objectives of the two stages of language modeling in pre-training and downstream hierarchy-based tasks, which is more feasible.
Technically, as is shown in Figure \ref{fig: HierICRF_Introduction}(b),
(1) Firstly, we construct a hierarchy-aware prompt to encourage the model to generate hierarchically repeated series.
(2) Secondly, this series will be fed into a verbalizer to obtain their masked language modeling (MLM) logits of labels in the hierarchical dependency tree.
(3) Finally, we use a hierarchical iterative CRF and initialize its transition matrix (e.g., transition scores between non-adjacent layers are set to a minimum to avoid erroneous cross-layer transfers) based on the hierarchical dependency tree to constrain the hierarchical dependency during the path routing process.
Combining these three stages, with the deepening of the hierarchically repeated reasoning process, the model can perform hierarchical consistency self-correction during each step to encourage predictions more accurate.
During the inference stage, we use the Viterbi algorithm~\cite{forney1973viterbi} to decode the series to obtain our final predictions.

The main contributions of this paper are summarized as:
\begin{itemize}
    \item To our best knowledge, we are among the first to investigate a few-shot HTC framework that emphasizes domain-hierarchy adaptation to bridge the gap between unstructured prior knowledge and downstream hierarchy.
    \item We proposed a unified framework that is suitable for any transformer-based architecture to efficiently mine prior knowledge within limited downstream labeled datasets for better few-shot learning. 
    \item We thoroughly study the hierarchical inconsistency problem.
    Experiments on BERT and T5 demonstrate that HierICRF outperforms the previous SOTA few-shot HTC methods on two popular datasets under extreme few-shot settings while achieving SOTA hierarchical consistency performance with an average of 9.3\% and 4.38\% CMacro-F1 improvements on WOS and DBpedia, respectively.
\end{itemize}

\begin{figure*}[t]
\centering
\includegraphics[width=0.95\textwidth]{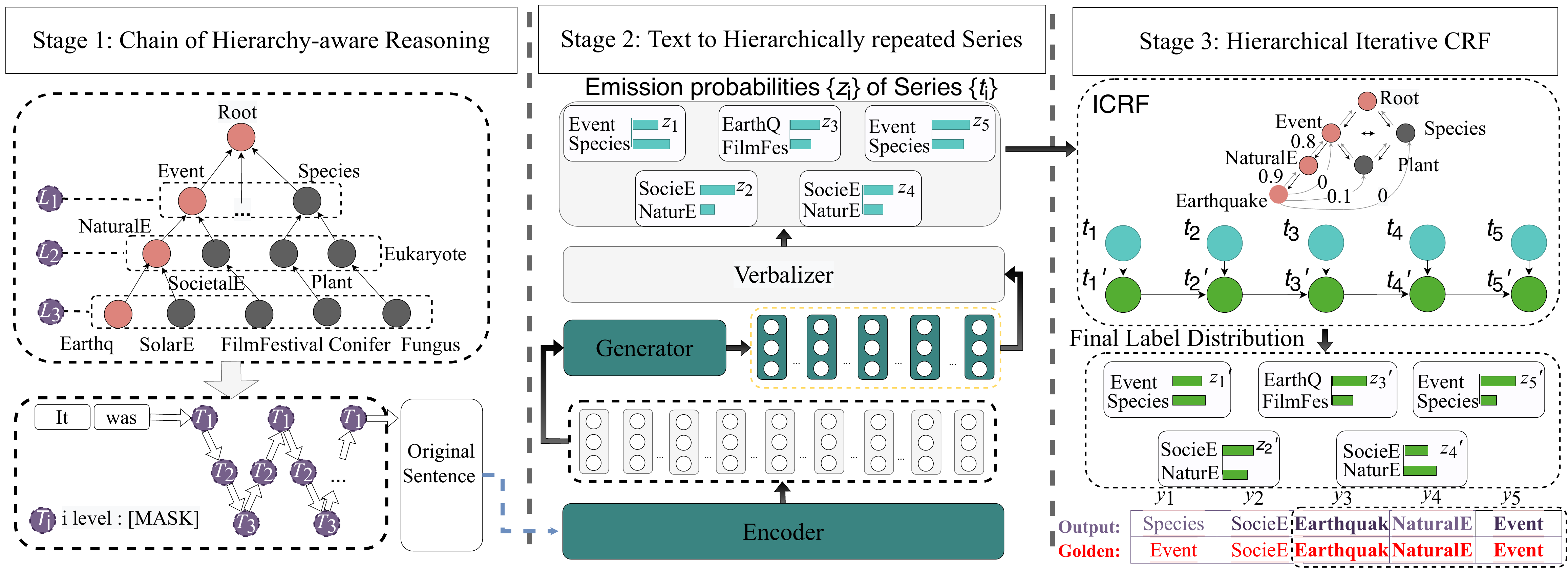}
\caption{
The overview of HierICRF. 
There are two ways to inject hierarchical constraints: (a) Chain of hierarchy-aware reasoning and (b) Hierarchical iterative CRF. 
At stage 3, we select the predictions $\{y_3, y_4, y_5\}$ (the last path routing iteration) as the final outputs.
}
\label{fig: HierICRF}
\end{figure*}

\section{Related Work}
\paragraph{Hierarchical Text Classification.} 
Current HTC research mainly focuses on how to incorporate hierarchical label knowledge to address imbalanced and large-scale label hierarchy challenges~\cite{mao-etal-2019-hierarchical}. 
Various approaches have been explored, including label-based attention modules~\cite{zhang2022hcn}, meta-learning~\cite{wu-etal-2019-learning}, and reinforcement learning methods~\cite{mao-etal-2019-hierarchical}. 
~\cite{zhou-etal-2020-hierarchy} proposes a more holistic approach named HiAGM by encoding the entire label structure with hierarchy encoders, which has shown greater performance improvements. 
Recent works~\cite{wang-etal-2021-concept,chen-etal-2021-hierarchy} have also explored matching learning and concept enrichment to exploit the relationship between text and label semantics. 
Later works such as HGCLR~\cite{wang-etal-2022-incorporating} and HPT~\cite{wang2022hpt} have migrated the label hierarchy into text encoding, achieving excellent performances through prompt tuning methods. 
\cite{ji2023hierarchical} proposes a multiple verbalizers framework to reduce the gap between PLMs and HTC for better few-shot learning.
Despite these advances, how to model HTC tasks with a unified hierarchy-aware paradigm is still underexplored, and there is a need to design a solution from the perspective of the path that performs well on consistency performance within limited training samples.

\section{Methodology}

\subsection{Problem Statements}
This paper mainly focuses on few-shot HTC tasks.
Different from previous works that assume a rich labeled dataset, we adopt the few-shot setting, i.e., only a limited number of samples are available for fine-tuning which is more practical as it assumes minimal resources.
We follow the greedy sampling algorithm proposed in~\cite{ji2023hierarchical} to obtain the support set~\textit{S} from the original HTC datasets, ensuring that each label path has exactly \textit{K}-shot samples.
Formally, the K-shot HTC task is defined as follows:
given a text \text{x} and a K-shot support set \textit{S} for the target mandatory-leaf path set $\textit{C}_\mathcal{T}$, the goal is to predict all golden paths on the label hierarchy tree, where $\textit{C}_\mathcal{T}$ consists of all paths from the root node to the leaf nodes on the label hierarchy tree $\mathcal{H}$.
In our settings, there is only one correct label for each layer.

\subsection{Framework Overview}
Figure~\ref{fig: HierICRF} shows the overall architecture of our proposed HierICRF.
Since the label structure that this paper focuses on is a tree, we view the HTC task as a process of a chain of thoughts via iterative CRF.
Specifically, in the first stage, we construct a hierarchy-aware reasoning chain $T_{chain}$ as our prompt to deepen the process of hierarchy-aware reasoning.
During the second stage, we use $T_{chain}$ to guide the generator to generate hierarchically repeated series $\textbf{\textit{t}}$, the $\textbf{\textit{t}}$ will be fed into a verbalizer to obtain their MLM logits $z$ in the label hierarchy tree.
Finally, during layer-to-layer transfer over the MLM logits $z_i$ of the series $\textbf{\textit{t}}_i$, we use a hierarchical iterative CRF (ICRF) to model its path routing by incorporating hierarchical information into the transition matrix of ICRF.

\subsection{Chain of Hierarchy-Aware Reasoning}
Due to the complex label dependency in HTC task, it is hard to adapt flat prior knowledge in PLMs to downstream hierarchical tasks.
Besides, hierarchical dependency information between labels at different layers or the same layers may be implicit.
To address the above issues, inspired by the chain of thought (COT)~\cite{wei2022chain} and math~\cite{zhu-etal-2023-solving} that solve mathematical or planning-related problems by decomposing complex problems into sub-processes, we propose a simple but effective method named hierarchy-aware reasoning chain to elegantly flattens HTC into a hierarchical loop-based language modeling process, thus allowing it to fully exploit the capabilities of PLM.
The details of obtaining the prompt of the hierarchy-aware reasoning chain named $T_{chain}$ are shown in Algorithm~\ref{alg: prefix}.
For example, when the depth of the hierarchy tree is 2 and the number of iterations $T_{chain}$ is 2, the reasoning chain template $T_{chain}$ is simply like "x. It was 1 level: [MASK] 2 level: [MASK] 1 level: [MASK] 2 level: [MASK] 1 level: [MASK]. ".
The $T_{chain}$ will later be used to guide the model to generate hierarchically repeated series. 

\subsection{Text to Hierarchically Repeated Series}
To verify the ability of our method for few-shot domain-hierarchy adaptation, we implement our method on both BERT and T5.
We first feed the input text $x$ wrapped with template $T_{chain}$ into the encoder to obtain the hidden states $\textbf{\textit{h}}_{1:n}$:
\begin{equation}
    \textbf{\textit{h}}_{1:n} = \text{Encoder}(T_{chain}(x)_{1:n})
\end{equation}
where $\textbf{\textit{h}}_{1:n} \in \mathbb{R}^{n \times r}$, and \textit{r} denotes the hidden state dimension of encoder and \textit{n} represents the length of $T_{chain}(x)$. 
We then obtain target hierarchically repeated series $\textbf{\textit{t}}_{1:l}=\{\textbf{\textit{t}}_i\}$ through:
\begin{equation}
    t_{1:l} = \text{Generator}(\textbf{\textit{h}}_{1:n})
\end{equation}
{\textit{For encoder-only LMs}}: Generator($\cdot$) means to directly extract a subset $\{\textbf{\textit{h}}_j\}$ consisting of hidden state vectors corresponding to all \texttt{[MASK]} tokens from $\textbf{\textit{h}}_{1:n}$ as our $\textbf{\textit{t}}_{1:l}$.
{\textit{For encoder-decoder LMs}}: Generator($\cdot$) represents feeding $\textbf{\textit{h}}_{1:n}$ into and prompt its decoder to obtain our final $\textbf{\textit{t}}_{1:l}$.

Furthermore, we construct a flat-verbalizer \textit{V} based on all labels on the hierarchical tree for label mapping learning.
The verbalizer is represented as a continuous vector $\textbf{\textit{W}}_V \in \mathbb{R}^{r \times m}$, where $m$ signifies the number of labels.
The embedding $\textbf{\textit{W}}_V$ is initialized by averaging the embeddings of its corresponding label tokens. 
We feed the series \{$\textbf{\textit{t}}_i$\} into the verbalizer to get the emission probabilities $z=\{z_1,...,z_l\}$. 

\begin{algorithm}[t]

\caption{Generate hierarchy-aware reasoning chain} 
\label{alg: prefix}
{\bf Input:} 
number of iterations $I_{chain}$, label hierarchy $\mathcal{H}$, depth of the $\mathcal{H}$ called $\mathcal{D}$ \\
{\bf Output:}
hierarchy reasoning chain
\begin{algorithmic}[1]
\STATE $T_{chain}$ $\rm  \gets\ ``It\ was"\ //Initialize\ the\ prompt$
\FOR{$i=1\ to\ |\mathcal{D}|$}
\STATE \rm $T_{chain}$ = $T_{chain}$ + ``i level: \texttt{[MASK]}"
\ENDFOR
\FOR{$i=\mathcal{D}-1\ to\ 1$}
\STATE \rm $T_{chain}$ = $T_{chain}$ + ``i level: \texttt{[MASK]}"
\ENDFOR
\FOR{$\_=1\ to\ |I_{chain}-1|$}
\FOR{$i=\mathcal{D}\ to\ 1$}
\STATE \rm $T_{chain}$ = $T_{chain}$ + ``i level: \texttt{[MASK]}"
\ENDFOR
\ENDFOR
\RETURN $T_{chain}$
\end{algorithmic}
\end{algorithm}

\subsection{Hierarchical Iterative Conditional Random Fields}
After obtaining the label distribution of each step in the hierarchy-aware reasoning chain, instead of directly classifying, we regard this hierarchically repeated series as a process of hierarchical path routing step-by-step.
Inspired by CRF~\cite{sutton2012introduction} widely used to model the transition of state space of time series in named entity recognition~\cite{nadeau2007survey}, here we model this sequencing process using hierarchical iterative CRF, injecting hierarchical constraint by optimizing transition matrices.

Formally, given a sentence \textit{x}, we use \textit{z} = \{$\textit{z}_1,...,\textit{z}_l$\} to represent a generic series where $z_i$ is the emission probability of the \textit{i}-th routing step.
$y = \{y_1,...,y_n\}$ represents the golden labels for \textit{z}.
The probabilistic model of a sequence CRF defines a set of conditional probabilities $p(y|z;\textbf{\textit{W}},b)$ over all possible label sequences \textit{y} given \textit{z} as:
\begin{equation}
    p(y|z;W,b) = \frac{\prod_{i=1}^n \psi_i(y_{i-1},y_i,z)}{\sum_{y^{'} \in \mathcal{H}} \prod_{i=1}^n \psi_i(y_{i-1}^{'},y_i^{'},z)}
\end{equation}
where $\psi_i(y',y,z) = \exp({\textbf{\textit{W}}_{y',y}^Tz_i + b_{y',y}}$) are potential functions, and $\textbf{\textit{W}}_{y',y}^T$ and $b_{y',y}$ denote the weight and bias corresponding to label pair $(y', y)$, respectively.
The final training objective is calculated as:
\begin{equation}
    L(\textbf{\textit{W}},b) = \sum_i log~p(y|z;\textbf{\textit{W}},b)
\end{equation}

Additionally, we initialize the transition scores between non-adjacent layers to a minimum to avoid erroneous cross-layer transfers. 
The transition score between Earthquake and Species in Figure~\ref{fig: HierICRF} is initialized to 0 before training.

\begin{table}[!b]
\normalsize

\begin{center}
\begin{tabular}{l|llll}
\toprule
\textbf{Datasets} & \textbf{DBpedia} &\textbf{WOS} \\ 
\midrule
Level 1 Categories & 9 & 7 \\
Level 2 Categories & 70 & 134\\
Level 3 Categories & 219 & NA\\
Number of documents & 381025 & 46985 \\
Mean document length &106.9 &200.7 \\ 
\bottomrule
\end{tabular}
\end{center}
\caption{Comparison of popular HTC datasets.}
\label{tab: dataset}
\end{table}

\begin{table*}[!t]
\renewcommand{\arraystretch}{0.8}
	\centering
	\small
	\setlength{\tabcolsep}{0.9mm}

	\begin{tabular}{lccccc}
        \toprule[1pt]
	       \multicolumn{1}{c}{\multirow{2}{*}{\begin{tabular}[c]{@{}c@{}} \\ K\end{tabular}}}
            & \multirow{2}{*}{\begin{tabular}[c]{@{}c@{}} \\ Method\end{tabular}} 
		 
            & \multicolumn{2}{c}{\textbf{WOS(Depth 2)}} 
            & \multicolumn{2}{c}{\textbf{DBpedia(Depth 3)}} 
         
         \\ \cmidrule {3-6} 
	     \multicolumn{1}{l}{} &  & Micro-F1 & Macro-F1 & Micro-F1  & Macro-F1  \\ 
      \midrule
		\multicolumn{1}{c}{\multirow{5}{*}{1}} 
  
		 & \multicolumn{1}{|l}{BERT (Vanilla FT)} 
            &2.99 $\pm$ 20.85 \textcolor[HTML]{387230}{(5.12)} 
            &0.16 $\pm$ 0.10 \textcolor[HTML]{387230}{(0.24)} 
            &14.43 $\pm$ 13.34 \textcolor[HTML]{387230}{(24.27)} 
            &0.29 $\pm$ 0.01 \textcolor[HTML]{387230}{(0.32)} 
            \\
		 & \multicolumn{1}{|l}{HiMatch-BERT~\cite{chen-etal-2021-hierarchy}} 
            &43.44 $\pm$ 8.90 \textcolor[HTML]{387230}{(48.26)}      
            &7.71 $\pm$ 4.90 \textcolor[HTML]{387230}{(9.32)}      
            &-  &-  
            \\
		 & \multicolumn{1}{|l}{HGCLR~\cite{wang-etal-2022-incorporating}} 
            &9.77 $\pm$ 11.77 \textcolor[HTML]{387230}{(16.32)}     
            &0.59 $\pm$ 0.10 \textcolor[HTML]{387230}{(0.63)}     
            &15.73 $\pm$ 31.07 \textcolor[HTML]{387230}{(25.13)}      
            &0.28 $\pm$ 0.10 \textcolor[HTML]{387230}{(0.31)}      
            \\
		 & \multicolumn{1}{|l}{HPT~\cite{wang2022hpt}} 
            &50.05 $\pm$ 6.80 \textcolor[HTML]{387230}{(50.96)}   
            &25.69 $\pm$ 3.31 \textcolor[HTML]{387230}{(27.76)} 
            &72.52 $\pm$ 10.20 \textcolor[HTML]{387230}{(73.47)}  
            &31.01 $\pm$ 2.61 \textcolor[HTML]{387230}{(32.50)} 
            \\
            & \multicolumn{1}{|l}{SoftVerb~\cite{schick-schutze-2021-exploiting}} 
         &{56.11} $\pm$ 7.44 \textcolor[HTML]{387230}{(58.13)}  
         &{41.35} $\pm$ 5.62 \textcolor[HTML]{387230}{(44.32)}   
         &{90.38} $\pm$ 0.10 \textcolor[HTML]{387230}{(90.89)} 
         &{82.72} $\pm$ 0.18 \textcolor[HTML]{387230}{(83.74)}
            \\ 
            & \multicolumn{1}{|l}{HierVerb~\cite{ji2023hierarchical}} 
         &{58.95} $\pm$ 6.38 \textcolor[HTML]{387230}{(61.76)}  
         &{44.96} $\pm$ 4.86 \textcolor[HTML]{387230}{(48.19)}   
         &{91.81} $\pm$ 0.07 \textcolor[HTML]{387230}{(91.95)} 
         &{85.32} $\pm$ 0.04 \textcolor[HTML]{387230}{(85.44)}
            \\ 
            \cmidrule {2-6}
         & \multicolumn{1}{|l}{HierICRF-BERT (Ours)} 
         &\textbf{59.40} $\pm$ \textbf{6.22} \textcolor[HTML]{387230}{(\textbf{62.01})}  
         &\underline{46.49} $\pm$ {3.91} \textcolor[HTML]{387230}{({49.54})}   
         &\textbf{92.05} $\pm$ \textbf{0.10} \textcolor[HTML]{387230}{(\textbf{92.11})}  
         &\underline{86.10} $\pm$ {0.10} \textcolor[HTML]{387230}{({86.78})}  
         \\ 
         & \multicolumn{1}{|l}{HierICRF-T5 (Ours)} 
        &\underline{59.20} $\pm$ {5.17}  \textcolor[HTML]{387230}{({61.97})}    
        &\textbf{47.72} $\pm$ \textbf{2.23}  \textcolor[HTML]{387230}{(\textbf{50.22})}    
        &\underline{91.94} $\pm$ {0.05}  \textcolor[HTML]{387230}{({92.01})}    
        &\textbf{86.70} $\pm$ \textbf{0.05}  \textcolor[HTML]{387230}{(\textbf{86.94})}   
        \\
         \midrule
        
         \multicolumn{1}{c}{\multirow{5}{*}{2}} 
         & \multicolumn{1}{|l}{BERT (Vanilla FT)} 
         &46.31 $\pm$ 0.65  \textcolor[HTML]{387230}{(46.85)}    
         &5.11 $\pm$ 1.31  \textcolor[HTML]{387230}{(5.51)}  
         &87.02 $\pm$ 3.89  \textcolor[HTML]{387230}{(88.20)}  
         &69.05 $\pm$ 26.81  \textcolor[HTML]{387230}{(73.28)}  
         \\
		 & \multicolumn{1}{|l}{HiMatch-BERT~\cite{chen-etal-2021-hierarchy}} 
         &46.41 $\pm$ 1.31  \textcolor[HTML]{387230}{(47.23)}    
         &18.97 $\pm$ 0.65  \textcolor[HTML]{387230}{(21.06)}  
         &-   &-  
         \\
	& \multicolumn{1}{|l}{HGCLR~\cite{wang-etal-2022-incorporating}} 
         &45.11 $\pm$ 5.02  \textcolor[HTML]{387230}{(47.56)}   
         &5.80 $\pm$ 11.63  \textcolor[HTML]{387230}{(9.63)}    
         &87.79 $\pm$ 0.40  \textcolor[HTML]{387230}{(88.42)}   
         &71.46 $\pm$ 0.17  \textcolor[HTML]{387230}{(71.78)}   
         \\
		 & \multicolumn{1}{|l}{HPT~\cite{wang2022hpt}} 
         &{57.45} $\pm$ 1.89 \textcolor[HTML]{387230}{(58.99)}    
         &{35.97} $\pm$ 11.89  \textcolor[HTML]{387230}{(39.94)}    
         &{90.32} $\pm$ 0.64  \textcolor[HTML]{387230}{(91.11)}    
         &{81.12} $\pm$ 1.33  \textcolor[HTML]{387230}{(82.42)}   
         \\
         & \multicolumn{1}{|l}{SoftVerb~\cite{schick-schutze-2021-exploiting}} 
         &{62.31} $\pm$ 13.24 \textcolor[HTML]{387230}{(65.02)}  
         &{49.33} $\pm$ 6.55 \textcolor[HTML]{387230}{(53.46)}   
         &{92.97} $\pm$ 0.20 \textcolor[HTML]{387230}{(93.04)} 
         &{87.61} $\pm$ 0.20 \textcolor[HTML]{387230}{(87.87)}
         \\ 
         & \multicolumn{1}{|l}{HierVerb~\cite{ji2023hierarchical}} 
         &\textbf{66.08} $\pm$ \textbf{4.19} \textcolor[HTML]{387230}{(\textbf{68.01})}  
         &{54.04} $\pm$ 3.24 \textcolor[HTML]{387230}{(56.69)}   
         &\underline{93.71} $\pm$ 0.01 \textcolor[HTML]{387230}{(93.87)}  
         &{88.96} $\pm$ 0.02 \textcolor[HTML]{387230}{(89.02)} 
         
         \\ \cmidrule {2-6}
         & \multicolumn{1}{|l}{HierICRF-BERT (Ours)} 
         &\underline{65.71} $\pm$ 3.69 \textcolor[HTML]{387230}{({67.87})}  
         &\underline{55.18} $\pm$ {3.11} \textcolor[HTML]{387230}{({57.12})}   
         &\textbf{94.22} $\pm$ \textbf{0.01} \textcolor[HTML]{387230}{(\textbf{94.22})}  
         &\textbf{89.31} $\pm$ \textbf{0.01} \textcolor[HTML]{387230}{({89.31})} 
         \\ 
         & \multicolumn{1}{|l}{HierICRF-T5 (Ours)} 
        &{65.52} $\pm$ {2.43}  \textcolor[HTML]{387230}{({66.97})}    
        &\textbf{56.11} $\pm$ \textbf{1.79}  \textcolor[HTML]{387230}{(\textbf{57.49})}    
        &{93.64} $\pm$ {0.01}  \textcolor[HTML]{387230}{({94.01})}    
        &\underline{89.22} $\pm$ {0.01}  \textcolor[HTML]{387230}{(\textbf{89.45})}   
        \\
         \midrule
        
        \multicolumn{1}{c}{\multirow{5}{*}{4}} 
         & \multicolumn{1}{|l}{BERT~(Vanilla FT)} 
         &56.00 $\pm$ 4.25  \textcolor[HTML]{387230}{(57.18)}    
         &31.04 $\pm$ 16.65  \textcolor[HTML]{387230}{(33.77)}   
         &92.94 $\pm$ 0.66  \textcolor[HTML]{387230}{(93.38)}   
         &84.63 $\pm$ 0.17  \textcolor[HTML]{387230}{(85.47)}     
         \\
		 & \multicolumn{1}{|l}{HiMatch-BERT~\cite{chen-etal-2021-hierarchy}} 
         &57.43 $\pm$ 0.01  \textcolor[HTML]{387230}{(57.43)}    
         &39.04 $\pm$ 0.01  \textcolor[HTML]{387230}{(39.04)}    
         &-   &-  
         \\
		 & \multicolumn{1}{|l}{HGCLR~\cite{wang-etal-2022-incorporating}} 
         &56.80 $\pm$ 4.24  \textcolor[HTML]{387230}{(57.96)}    
         &32.34 $\pm$ 15.39  \textcolor[HTML]{387230}{(33.76)}    
         &93.14 $\pm$ 0.01  \textcolor[HTML]{387230}{(93.22)}   
         &84.74 $\pm$ 0.11  \textcolor[HTML]{387230}{(85.11)}   
         \\
		 & \multicolumn{1}{|l}{HPT~\cite{wang2022hpt}} 
         &{65.57} $\pm$ 1.69  \textcolor[HTML]{387230}{(67.06)}    
         &{45.89} $\pm$ 9.78  \textcolor[HTML]{387230}{(49.42)}     
         &{94.34} $\pm$ 0.28  \textcolor[HTML]{387230}{(94.83)}    
         &{90.09} $\pm$ 0.87  \textcolor[HTML]{387230}{(91.12)}   
         \\
         & \multicolumn{1}{|l}{SoftVerb~\cite{schick-schutze-2021-exploiting}}  
         &{69.58} $\pm$ {3.27}  \textcolor[HTML]{387230}{({71.00})}      
         &{58.53} $\pm$ {1.64}  \textcolor[HTML]{387230}{({60.18})}      
         &{94.47} $\pm$ {0.10}  \textcolor[HTML]{387230}{({94.74})}      
         &{90.25} $\pm$ {0.10}  \textcolor[HTML]{387230}{({90.73})}  
         \\ 
         & \multicolumn{1}{|l}{HierVerb~\cite{ji2023hierarchical}}  
         &{72.58} $\pm$ {0.83}  \textcolor[HTML]{387230}{({73.64})}      
         &{63.12} $\pm$ {1.48}  \textcolor[HTML]{387230}{({64.47})}      
         &\underline{94.75} $\pm$ {0.13}  \textcolor[HTML]{387230}{({95.13})}      
         &{90.77} $\pm$ {0.33}  \textcolor[HTML]{387230}{({91.43})}  
         \\ 
         \cmidrule {2-6}
         & \multicolumn{1}{|l}{HierICRF-BERT (Ours)}  
         &\textbf{73.83} $\pm$ \textbf{0.71}  \textcolor[HTML]{387230}{(\textbf{74.19})}      
         &\underline{65.40} $\pm$ {0.69}  \textcolor[HTML]{387230}{({65.86})}      
         &\textbf{95.14} $\pm$ \textbf{0.15}  \textcolor[HTML]{387230}{(\textbf{95.20})}      
         &\textbf{91.20} $\pm$ \textbf{0.05}  \textcolor[HTML]{387230}{(\textbf{91.81})}        
         \\
         & \multicolumn{1}{|l}{HierICRF-T5 (Ours)} 
        &\underline{73.22} $\pm$ {0.36}  \textcolor[HTML]{387230}{({60.11})}    
        &\textbf{65.61} $\pm$ \textbf{0.54}  \textcolor[HTML]{387230}{(\textbf{66.12})}    
        &{94.66} $\pm$ {0.01}  \textcolor[HTML]{387230}{({95.10})}    
        &\underline{90.89} $\pm$ {0.01}  \textcolor[HTML]{387230}{({91.33})}
        \\
         \midrule
        
        \multicolumn{1}{c}{\multirow{5}{*}{8}} 
         & \multicolumn{1}{|l}{BERT~(Vanilla FT)} 
         &66.24 $\pm$ 1.96  \textcolor[HTML]{387230}{(67.53)}    
         &50.21 $\pm$ 5.05  \textcolor[HTML]{387230}{(52.60)}    
         &94.39 $\pm$ 0.06  \textcolor[HTML]{387230}{(94.57)}   
         &87.63 $\pm$ 0.28  \textcolor[HTML]{387230}{(87.78)}   
         \\
		 & \multicolumn{1}{|l}{HiMatch-BERT~\cite{chen-etal-2021-hierarchy}} 
         &69.92 $\pm$ 0.01  \textcolor[HTML]{387230}{(70.23)}    
         &57.47 $\pm$ 0.01  \textcolor[HTML]{387230}{(57.78)}  
         &-   &-  
         \\
		 & \multicolumn{1}{|l}{HGCLR \cite{wang-etal-2022-incorporating}} 
         &68.34 $\pm$ 0.96  \textcolor[HTML]{387230}{(69.22)}    
         &54.41 $\pm$ 2.97  \textcolor[HTML]{387230}{(55.99)}    
         &94.70 $\pm$ 0.05  \textcolor[HTML]{387230}{(94.94)}   
         &88.04 $\pm$ 0.25  \textcolor[HTML]{387230}{(88.61)}    
         \\
		 & \multicolumn{1}{|l}{HPT~\cite{wang2022hpt}} 
         &{76.22} $\pm$ 0.99  \textcolor[HTML]{387230}{(77.23)}    
         &{67.20} $\pm$ 1.89  \textcolor[HTML]{387230}{(68.63)}   
         &{95.49} $\pm$ 0.01  \textcolor[HTML]{387230}{(95.57)}    
         &{92.35} $\pm$ 0.03  \textcolor[HTML]{387230}{(92.52)}    
         \\
         & \multicolumn{1}{|l}{SoftVerb~\cite{schick-schutze-2021-exploiting}}  
         &{75.99} $\pm$ {0.47}  \textcolor[HTML]{387230}{({76.77})}      
         &{66.99} $\pm$ {0.27}  \textcolor[HTML]{387230}{({67.50})}      
         &{95.48} $\pm$ {0.01}  \textcolor[HTML]{387230}{({95.64})}      
         &{92.06} $\pm$ {0.01}  \textcolor[HTML]{387230}{({92.37})}  
         \\ 
         & \multicolumn{1}{|l}{HierVerb~\cite{ji2023hierarchical}} 
        &\underline{78.12} $\pm$ {0.55}  \textcolor[HTML]{387230}{(\textbf{78.87})}    
        &{69.98} $\pm$ {0.91}  \textcolor[HTML]{387230}{({71.04})}    
        &\underline{95.69} $\pm$ {0.01}  \textcolor[HTML]{387230}{({95.70})}    
        &{92.44} $\pm$ {0.01}  \textcolor[HTML]{387230}{({92.51})}
         \\ 
         \cmidrule {2-6}
         & \multicolumn{1}{|l}{HierICRF-BERT (Ours)} 
        &\textbf{78.54} $\pm$ \textbf{0.25}  \textcolor[HTML]{387230}{({78.69})}    
        &\underline{70.79} $\pm$ {0.38}  \textcolor[HTML]{387230}{({71.35})}    
        &\textbf{95.80} $\pm$ \textbf{0.01}  \textcolor[HTML]{387230}{(\textbf{95.85})}    
        &\textbf{92.77} $\pm$ \textbf{0.01}  \textcolor[HTML]{387230}{({92.82})}       
        \\ 
        & \multicolumn{1}{|l}{HierICRF-T5 (Ours)} 
        &{77.78} $\pm$ {0.17}  \textcolor[HTML]{387230}{({78.64})}    
        &\textbf{71.62} $\pm$ \textbf{0.10}  \textcolor[HTML]{387230}{(\textbf{71.90})}    
        &{95.55} $\pm$ {0.01}  \textcolor[HTML]{387230}{({95.70})}    
        &\underline{92.69} $\pm$ {0.01}  \textcolor[HTML]{387230}{({92.79})}
        \\
        \midrule
        
        \multicolumn{1}{c}{\multirow{5}{*}{16}} 
         & \multicolumn{1}{|l}{BERT~(Vanilla FT)} 
         &75.52 $\pm$ 0.32  \textcolor[HTML]{387230}{(76.07)}    
         &65.85 $\pm$ 1.28  \textcolor[HTML]{387230}{(66.96)}    
         &95.31 $\pm$ 0.01  \textcolor[HTML]{387230}{(95.37)}   
         &89.16 $\pm$ 0.07  \textcolor[HTML]{387230}{(89.35)}   
         \\
		 & \multicolumn{1}{|l}{HiMatch-BERT~\cite{chen-etal-2021-hierarchy}} 
         &77.67 $\pm$ 0.01  \textcolor[HTML]{387230}{(78.24)}    
         &68.70 $\pm$ 0.01  \textcolor[HTML]{387230}{(69.58)}  
         &-   
         &-  
         \\
		 & \multicolumn{1}{|l}{HGCLR~\cite{wang-etal-2022-incorporating}} 
         &76.93 $\pm$ 0.52  \textcolor[HTML]{387230}{(77.46)}  
         &67.92 $\pm$ 1.21  \textcolor[HTML]{387230}{(68.66)}  
         &95.49 $\pm$ 0.04  \textcolor[HTML]{387230}{(95.63)}    
         &89.41 $\pm$ 0.09  \textcolor[HTML]{387230}{(89.71)}    
         \\
		 & \multicolumn{1}{|l}{HPT~\cite{wang2022hpt}} 
         &{79.85} $\pm$ 0.41  \textcolor[HTML]{387230}{(80.58)}    
         &{72.02} $\pm$ 1.40  \textcolor[HTML]{387230}{(73.31)}    
         &{96.13} $\pm$ 0.01  \textcolor[HTML]{387230}{(96.21)}    
         &{93.34} $\pm$ {0.02}  \textcolor[HTML]{387230}{({93.45})}   
         \\
         & \multicolumn{1}{|l}{SoftVerb~\cite{schick-schutze-2021-exploiting}}  
         &{79.62} $\pm$ {0.85}  \textcolor[HTML]{387230}{({80.68})}      
         &{70.95} $\pm$ {0.62}  \textcolor[HTML]{387230}{({71.84})}      
         &{95.94} $\pm$ {0.15}  \textcolor[HTML]{387230}{({96.18})}      
         &{92.89} $\pm$ {0.20}  \textcolor[HTML]{387230}{({93.37})}  
         \\ 
         & \multicolumn{1}{|l}{HierVerb~\cite{ji2023hierarchical}} 
         &{80.93} $\pm$ {0.10}  \textcolor[HTML]{387230}{({\textbf{81.26}})}      
         &{73.80} $\pm$ {0.12}  \textcolor[HTML]{387230}{({{74.19}})}      
         &\underline{96.17} $\pm$ {0.01}  \textcolor[HTML]{387230}{({96.21})}      
         &{93.28} $\pm$ 0.06  \textcolor[HTML]{387230}{(93.49)}     
            
         \\ \cmidrule {2-6}
         & \multicolumn{1}{|l}{HierICRF-BERT (Ours)} 
         &\textbf{81.02} $\pm$ \textbf{0.10}  \textcolor[HTML]{387230}{({81.20})}      
         &\underline{74.05} $\pm$ {0.10}  \textcolor[HTML]{387230}{({74.15})}      
         &\textbf{96.22} $\pm$ \textbf{0.01}  \textcolor[HTML]{387230}{(\textbf{96.25})}      
         &\underline{93.38} $\pm$ {0.02}  \textcolor[HTML]{387230}{({93.60})}      
         \\ 
        & \multicolumn{1}{|l}{HierICRF-T5 (Ours)} 
        &\underline{80.94} $\pm$ {0.05}  \textcolor[HTML]{387230}{({81.05})}    
        &\textbf{75.23} $\pm$ \textbf{0.05}  \textcolor[HTML]{387230}{(\textbf{75.88})}    
        &{96.11} $\pm$ {0.01}  \textcolor[HTML]{387230}{({95.85})}    
        &\textbf{93.56} $\pm$ \textbf{0.01}  \textcolor[HTML]{387230}{(\textbf{93.70})}
        \\
        \midrule
	\end{tabular}
\caption{Results of 1/2/4/8/16-shot HTC. F1 scores on WOS and DBpedia. 
        We report the mean F1 scores ($\%$) over 3 random seeds. 
        Bold: best results. 
        Underlined: second highest. 
        For baseline models, we report the F1 scores from their original paper.
        }
\label{tab:main_results}
\end{table*}

\subsection{Decoding}
After training, the decoding is to search for the label sequence $y^*$ with the maximum conditional probability:
\begin{equation}
    y^* = \argmax~p(y|x;\textbf{\textit{W}},b);y\in \mathcal{H}
\end{equation}
where we pick out the outputs of the last path routing iteration from $y^*$  as our final predictions.

\section{Experiments}

\subsection{Experimental Settings}

\paragraph{Datasets and Evaluation Metrics.}
We evaluate our method and all baselines on two popular datasets for HTC: Web-of-Science (WOS)~\cite{kowsari2017hdltex}, DBpedia~\cite{sinha-etal-2018-hierarchical}.
WOS is a database that includes abstracts of published papers, among other bibliographic information such as author names, journal titles, and publication dates.
DBpedia is a bigger dataset with labels from Wikipedia meta information provider DBpedia with a three-level hierarchy.
Table~\ref{tab: dataset} presents the statistical details. 
There are differences in the domain distribution of these two datasets.

We measure the experimental results with Macro-F1 and Micro-F1.
To more thoroughly assess the hierarchical consistency, we utilize the path-constrained C-metric proposed in~\cite{yu2022constrained} and the path-based P-metric proposed in~\cite{ji2023hierarchical}.
The C-metric only considers a correct prediction for a label node to be valid if all of its ancestor nodes are predicted correctly. 
In contrast, the P-metric requires that all of the ancestors and child nodes on the path to which the label node belongs are predicted correctly for its correct prediction to be considered valid.

\paragraph{Baselines.}
For performance comparison of various models from a different perspective, we select the following strong baselines:
~{HiMatch-BERT}~\cite{chen-etal-2021-hierarchy},
~{HGCLR}~\cite{wang-etal-2022-incorporating},
~{HPT}~\cite{wang2022hpt} and
~{HierVerb}~\cite{ji2023hierarchical}.
We also perform {Vanilla Fine-Tuning (Vanilla FT)}~\cite{devlin-etal-2019-bert} and {Vanilla Soft Verbalizer (SoftVerb)}~\cite{schick-schutze-2021-exploiting} method on the Few-shot HTC task.
Vanilla FT is a simple method consisting of a Binary CrossEntropy loss for ordinary multi-label classification followed by a classifier.
SoftVerb uses the traditional template "x. It was 1:level [MASK] 2:level [MASK].", then the hidden states of all positions are fed into the verbalizer to obtain label logits.
Note that SoftVerb, HPT, and HierVerb are prompt-based methods and all baselines above are limited to encoder-only architectures. 
Considering that the current research based on encoder-decoder performs poorly under few-shot scenario, we construct a strong baseline called SoftVerb-T5 by removing all components on HierICRF-T5 in the ablation experiment for a fair comparison.

\paragraph{Backbone and Implementation Details.}
We adopt both BERT~\texttt{(BERT-base-uncased)}~\cite{devlin-etal-2019-bert} and T5~\texttt{(T5-base)}~\cite{raffel2020exploring} as the main backbone of our experiments, for additional experiments we use BERT~(\texttt{BERT-base-uncased}) by default.

The batch size is 8.
We use a learning rate of 5e-5 for BERT and 3e-5 for T5 and train the model for 20 epochs. 
For soft verbalizer and ICRF, we use a learning rate of 1e-4 to encourage a faster convergence.
After each epoch, we evaluate the model's performance on the development set and set early stopping to 5 as usual.
For the baseline models, we follow the hyperparameter set as specified in their respective papers.

\begin{table*}[!t]
\renewcommand{\arraystretch}{0.8}
	\centering
        
	\footnotesize
	\setlength{\tabcolsep}{1.5mm}
	\begin{tabular}{clcccccccc}
		\toprule[1pt]
		\multicolumn{1}{c}{\multirow{2}{*}{\begin{tabular}[c]{@{}c@{}} \\ 
            K
        \end{tabular}}} & \multicolumn{1}{l}{\multirow{2}{*}{\begin{tabular}[c]{@{}c@{}} \\ 
            Method\end{tabular}}}
		& \multicolumn{4}{c}{\textbf{WOS}} & \multicolumn{4}{c}{\textbf{DBpedia}} \\ \cmidrule {3-10} 
		& & PMicro-F1 & PMacro-F1 & CMicro-F1 & CMacro-F1 & PMicro-F1 & PMacro-F1 & CMicro-F1 & CMacro-F1 \\ \midrule
  
            \multicolumn{1}{c}{\multirow{5}{*}{1}} 
		& \multicolumn{1}{|l}{HierICRF-BERT} &\textbf{42.63}  &\textbf{41.62} &\textbf{57.35} &\textbf{44.06}   &\textbf{84.95} &\textbf{79.44} &\textbf{90.52} &\textbf{84.42}\\
            & \multicolumn{1}{|l}{HierVerb} &{39.77}  &{37.24} &{55.18} &{39.42} &{83.56}  &{77.96} &{89.80} &{81.78} \\
		& \multicolumn{1}{|l}{HPT} &19.97  &17.47 &49.10 &22.92 &61.08  &57.80 &82.84 &66.99  \\
		& \multicolumn{1}{|l}{HGCLR} &0.0  &0.0 &2.21 &0.09 &0.0  &0.0 &28.05 &0.24  \\
		& \multicolumn{1}{|l}{Vanilla FT} &0.0  &0.0  &0.96 &0.04 &0.0  &0.0  &28.08 &0.24 \\
		\midrule
		
		\multicolumn{1}{c}{\multirow{5}{*}{2}} 
		& \multicolumn{1}{|l}{HierICRF} &\textbf{52.62} &\textbf{49.39} &\textbf{64.07} &\textbf{52.64} &\textbf{90.14} &\textbf{87.66} &\textbf{94.54} &\textbf{90.26} \\
            & \multicolumn{1}{|l}{HierVerb} &{50.15}  &{47.98} &{62.90} &{49.67} &{88.58}  &{86.35} &{93.61} &{88.96} \\
		& \multicolumn{1}{|l}{HPT} &28.27  &26.51 &56.64 &33.50 &82.36  &81.41 &92.31 &86.43 \\
		& \multicolumn{1}{|l}{HGCLR} &1.39 &1.49 &45.01 &4.88 &54.55  &3.72 &67.70 &26.41  \\
		& \multicolumn{1}{|l}{Vanilla FT} &1.43 &1.42 &45.75 &4.95 &53.83  &3.71 &67.72 &26.89  \\
		\midrule
		
		\multicolumn{1}{c}{\multirow{5}{*}{4}} 
		& \multicolumn{1}{|l}{HierICRF} &\textbf{63.91}  &\textbf{60.17} &\textbf{73.45}  &\textbf{62.88} &\textbf{93.20}  &\textbf{92.51} &\textbf{95.88}  &\textbf{93.31}\\
            & \multicolumn{1}{|l}{HierVerb} &{62.16}  &{59.70} &{72.41} &{61.19} &{91.90}  &{91.38} &{95.74} &{92.87} \\
		& \multicolumn{1}{|l}{HPT} &50.96  &48.76 &69.43 &55.27 &87.61  &87.04 &94.50 &90.42 \\
		& \multicolumn{1}{|l}{HGCLR} &29.94  &27.70  &57.43  &34.03 &55.34  &3.76  &67.54  &28.60 \\
		& \multicolumn{1}{|l}{Vanilla FT} &22.97  &20.73  &55.10 &27.50 &55.15  &3.74  &67.44 &28.32 \\

		\bottomrule[1pt]
	\end{tabular}
\caption
	{
    	Consistency experiments on the WOS and DBpedia datasets using two path-constraint metrics.
            PMicro-F1 and PMacro-F1 are our proposed path-based consistency evaluation P-metric.
            We report the mean F1 scores (\%) over 3 random seeds.
            All experiments use their respective metrics as a signal for early stopping.
	}
\label{tab: consistency_complete}
\end{table*}

\subsection{Main Results}
\label{sec: main_results}
Main experimental results are reported in Table~\ref{tab:main_results}.
By optimizing from a path routing perspective, HierICRF outperforms overall comparison baselines under most situations.

We first find out that prompt-based methods outperform vanilla FT by a dramatic margin in the case of no more than 4-shot settings.
On average, 67.01\%, 13.30\%, and 10.01\% Micro-F1 improvements are achieved by HierICRF-BERT compared to the vanilla FT.
However, although HPT is designed as a prompt-based method, its few-shot results are not satisfactory.
The reason is obvious since the overfitting problem of the GNN layers becomes serious when labeled data used for fine-tuning is limited.

Second, HierICRF-BERT achieves 1.53\%, 1.14\%, and 2.28\% Macro-F1 improvements from the best baselines on 1, 2, and 4-shot on WOS, respectively.
Besides, HierICRF-T5 has a substantial improvement with an average Macro-F1 of 2.08\% on WOS and 0.46\% on DBpedia compared to the previous SOTA HierVerb while keeping competitive with HierICRF-BERT on Micro-F1.

Additionally, it is clear that both the HierICRF-BERT and HierICRF-T5's Micro-F1 and Macro-F1 change very slightly from 1 to 16 shots on DBpedia while other models except HierVerb are particularly dependent on the increase of labeled training samples.
For example, as the shots become fewer, the HierICRF-BERT's Micro-F1 changes from 96.22\% to 92.05\% while HGCLR's Micro-F1 changes from 95.49\% to 15.73\%.
The results indicate that our method can efficiently mine the prior knowledge in pre-training for hierarchical tasks in the case of extremely few samples instead of relying too much on limited training samples to optimize the model.

\begin{table}[!t]
\renewcommand{\arraystretch}{0.8}
	\centering
        \setlength\tabcolsep{3pt} 
        
	\small
	\setlength{\tabcolsep}{0.4mm}
	\begin{tabular}{clcccc}
		\toprule[1pt]
		\multicolumn{1}{c}{\multirow{2}{*}{\begin{tabular}[c]{@{}c@{}} \\ K\end{tabular}}}
		 & \multicolumn{1}{l}{\multirow{2}{*}{\begin{tabular}[c]{@{}c@{}} \\ Ablation Models\end{tabular}}}
		& \multicolumn{2}{c}{\textbf{BERT}} & \multicolumn{2}{c}{\textbf{T5}} \\ 
            \cmidrule {3-6} 
		& & Micro-F1 & Macro-F1 & Micro-F1 & Macro-F1  \\ \midrule
		
		\multicolumn{1}{c}{\multirow{4}{*}{1}}
		& \multicolumn{1}{|l}{Ours} &\textbf{59.40}  &\textbf{46.49}  &\textbf{59.20} &\textbf{47.72} \\
		& \multicolumn{1}{|l}{$r.m.$ ICRF loss} &57.54  &43.25  &57.82 &45.77 \\
		& \multicolumn{1}{|l}{$r.m.$ CHR} &57.82  &43.71 &56.88 &44.25  \\
		& \multicolumn{1}{|l}{$r.m.$ ICRF\&CHR} &56.11  &41.35  &57.16 &44.83 \\
		\midrule
		\multicolumn{1}{c}{\multirow{4}{*}{2}} 
		& \multicolumn{1}{|l}{Ours} &\textbf{65.71}  &\textbf{55.18} &\textbf{65.52} &\textbf{56.11} \\
		& \multicolumn{1}{|l}{$r.m.$ ICRF loss} &64.11  &52.05 &64.29 &53.47 \\
		& \multicolumn{1}{|l}{$r.m.$ CHR} &64.52  &52.39 &64.73 &54.82 \\
		& \multicolumn{1}{|l}{$r.m.$ ICRF\&CHR} &62.31  &49.33  &62.07 &49.67 \\
		\midrule
		
		\multicolumn{1}{c}{\multirow{4}{*}{4}} 
		& \multicolumn{1}{|l}{Ours} &\textbf{73.83}  &\textbf{65.40} &\textbf{73.22} &\textbf{65.61} \\
		& \multicolumn{1}{|l}{$r.m.$ ICRF loss} &71.78  &63.99 &71.21 &63.48 \\
		& \multicolumn{1}{|l}{$r.m.$ CHR} &71.51  &63.29 &72.52 &64.15\\
		& \multicolumn{1}{|l}{$r.m.$ ICRF\&CHR} &69.58  &58.83 &70.24 &59.75  \\
		\midrule
		
		\multicolumn{1}{c}{\multirow{4}{*}{8}} 
		& \multicolumn{1}{|l}{Ours} &\textbf{78.54}  &\textbf{70.79} &\textbf{77.78} &\textbf{71.62} \\
		& \multicolumn{1}{|l}{$r.m.$ ICRF loss} &76.44  &68.23 &76.52 &69.30 \\
		& \multicolumn{1}{|l}{$r.m.$ CHR} &77.29  &69.60 &77.63 &70.44 \\
		& \multicolumn{1}{|l}{$r.m.$ ICRF\&CHR} &75.99  &66.99 &75.24 &67.11  \\
		\midrule
		
		\multicolumn{1}{c}{\multirow{4}{*}{16}} 
		& \multicolumn{1}{|l}{Ours} &\textbf{81.02}  &\textbf{74.05} &\textbf{80.94} &\textbf{75.23} \\
		& \multicolumn{1}{|l}{$r.m.$ ICRF loss} &80.04  &72.71 &80.15 &72.41 \\
		& \multicolumn{1}{|l}{$r.m.$ CHR} &80.54  &73.64 &80.46 &73.84\\
		& \multicolumn{1}{|l}{$r.m.$ ICRF\&CHR} &79.62  &70.95 &{78.91} &{71.74}  \\
		
		\bottomrule[1pt]
	\end{tabular}
\caption
	{
    	Ablation experiments on WOS. 
    	$r.m.$ stands for $remove$.
    	We report the mean F1 scores ($\%$) over 3 random seeds. 
            ICRF stands for Hierarchical Iterative CRF while CHR stands for Chain of Hierarchy-aware Reasoning.
            Note when all mechanisms are removed, HierICRF is equivalent to Vanilla SoftVerb.
	}
\label{tab: ablation}
\end{table}

\subsection{Hierarchical Consistency Performance}
Table~\ref{tab: consistency_complete} further studies the consistency performance.
Our method still maintains SOTA consistency performance in the absence of labeled training corpora.
It is clear that HGCLR and BERT (Vanilla FT) which uses the direct fitting method only achieve 0 points in PMicro-F1 and PMacro-F1 under the 1-shot setting.
Compared with the best-performing baseline HierVerb, on average, HierICRF-BERT outperforms 7.08\% and 4.25\% PMicro-F1 scores, and 6.26\% and 3.92\% PMacro-F1, and 4.38\% and 1.79\% CMicro-F1, and 9.3\% and 4.38\% CMacro-F1 scores on WOS, DBpedia, respectively.
As for HPT and HierVerb, directly extra embedding injection to the pre-trained LM pays less attention to the hierarchy consistency.
The results highlight that the proposed paradigm allows model to optimize from a path routing perspective, more consideration is given to the label dependency during the process of iteratively transiting between layers on the hierarchically repeated series to better deal with the hierarchical inconsistency problem.
Surprisingly, when more training data are given, the performance gap between HierICRF and all other baselines gradually decreases as we may hypothesize.
We conjecture that it is because although all baseline methods lack domain-hierarchy adaptation in the paradigm, data-hungry based methods such as GNN can still learn hierarchical consistency dependency through directly overfitting, thereby gradually improving consistency performance.

\subsection{Ablation Experiments}
To illustrate the effect of our proposed mechanisms, we conduct ablation studies on WOS, as shown in Table~\ref{tab: ablation}.
When both ICRF and CHR are removed, HierICRF is equivalent to Vanilla SoftVerb.
Removing ICRF results in significant performance degradation, with an average of 1.72\% Micro-F1 and 2.38\% Macro-F1 on BERT and 1.72\% Micro-F1 and 2.37\% Macro-F1 on T5, meaning that ICRF plays an important role in incorporating hierarchy dependencies.
Furthermore, the performance decreases even more when both ICRF and CHR are removed, e.g., in the 4-shot case, Macro-F1 even drops by 6.57\% compared to the full method deployed on BERT.
This firmly highlights the significance of combining chain of hierarchy-aware reasoning and hierarchical iterative CRF allows HierICRF to gradually inject label hierarchy information by feeding back the accumulated error transition of label nodes in the process of path routing step-by-step.

\subsection{Benefit in a Full-Shot Setup}
We conduct experiments using a full-shot setting and use the hyperparameter set directly from the few-shot settings. 
For the baseline models, we reproduce their experiments based on the settings in their original paper. 
While the main focus of our work is on the performance under few-shot settings, it is worth noting that HierICRF surprisingly outperforms all baselines, such as HGCLR, HPT, and HierVerb, in the full-shot setting.
As shown in Table~\ref{tab:full_shot}, our Micro-F1 is slightly higher than HGCLR, HPT, and HierVerb, while Macro-F1 significantly beats them by 0.28\% on average.
Besides, HierICRF significantly outperforms BERT (Vanilla FT), HiMatch, and SoftVerb.
\begin{table}[!t]
\renewcommand{\arraystretch}{0.8}
    \centering
    
    \small
    \begin{tabular}{l|cc|cc}
    \toprule
    & \multicolumn{2}{|c|}{\textbf{WOS}}
    & \\
    \textbf{Methods}& Micro-F1 & Macro-F1 \\
    \midrule
        HierICRF &\textbf{87.12}  &\textbf{81.65}\\
        HierVerb &87.00  &81.57 \\
        HPT  &87.10  &81.44 \\
        SoftVerb  &86.80  &81.23 \\
        HGCLR &87.08  &81.11 \\
        HiMatch-BERT &86.70  &81.06 \\
        BERT (Vanilla FT) &85.63  &79.07 \\
    \bottomrule
    \end{tabular}
\caption{Full-shot experiments using \texttt{BERT-base-uncased} on WOS dataest .
     }
\label{tab:full_shot}
\end{table}

\subsection{Effect of Reasoning Chain’s Length}
To further study the effects of the hierarchy-aware reasoning chain, we conduct experiments on the iteration length of the reasoning chain.
As we can see in Table~\ref{tab: iterative_chain}, the performances on both WOS and DBpedia degrade substantially as $I_{chain}$ decreases,  with an average Micro-F1 of 1.82\% and 0.54\% on WOS and DBpedia.
The results show that our method can correct more hierarchical inconsistencies between time steps and feedback to model optimization when performing hierarchical transitions in longer hierarchically repeated series.

\begin{table}[!t]
\renewcommand{\arraystretch}{0.8}
    \centering
    \small
    
    \begin{tabular}{c|cc|cc}
    \toprule
    & \multicolumn{2}{c|}{{WOS}} 
    & \multicolumn{2}{|c}{{DBpedia}}
     \\
    \textbf{$I_{chain}$}& Micro-F1 & Macro-F1 & Micro-F1 & Macro-F1 
    \\
    \midrule
    5  &\textbf{73.83}  &\textbf{65.40} &\textbf{95.14} &\textbf{91.20} \\
    3  &72.86  &65.01 &95.05 &91.11 \\
    1  &72.16  &64.55 &94.74 &90.51 \\
    0  &72.01  &64.29 &94.60 & 90.37 \\
    \bottomrule
    \end{tabular}
\caption{Effect of Hierarchy-aware Chain’s Length.
            Here we conduct experiments under 4-shot settings.
            }
\label{tab: iterative_chain}
\end{table}

\subsection{Effect of Model Scales}
To further study the ability of HierICRF to utilize the prior knowledge of the PLMs, we conduct experiments on \texttt{BERT-large-uncased} and \texttt{T5-large}.
Table~\ref{tab: model_scale} demonstrates that HierICRF consistently outperforms all baseline models in all shot settings.
We find that the gap is even significantly larger for HierICRF and all other baseline models compared to using \texttt{BERT-base-uncased}.
For example, compared with HierVerb, HierICRF-BERT(\texttt{BERT-large-uncased}) achieves a 1.53\% Macro-F1 and a 0.45\% Micro-F1 scores increase under 1-shot setting.
But the improvements of Macro-F1 and Micro-F1 are 1.81\% and 0.81\% under \texttt{BERT-base-uncased}, respectively.
Surprisingly, when using \texttt{T5-large}, we find that its performance improvement relative to \texttt{T5-base} is greater than that of the improvement obtained by all all BERT-based methods (including HierICRF-BERT) from \texttt{BERT-base-uncased} to \texttt{BERT-large-uncased}, and even achieve SOTA under no more than 2-shot settings.
The findings further underscore that HierICRF outperforms all baseline models in effectively leveraging the prior knowledge embedded within larger language models. This advantage becomes even more pronounced as the scale of the language model increases, highlighting the significant impact of HierICRF's ability to harness this expansive prior knowledge.

\begin{table}[!t]
\renewcommand{\arraystretch}{0.8}
	\centering
        
	\footnotesize
	\setlength{\tabcolsep}{0.4mm}
	\begin{tabular}{clcc}
		\toprule[1pt]
		\multicolumn{1}{c}{\multirow{2}{*}{\begin{tabular}[c]{@{}c@{}} \\ K\end{tabular}}}
		 & \multicolumn{1}{l}{\multirow{2}{*}{\begin{tabular}[c]{@{}c@{}} \\ Method\end{tabular}}}
		& \multicolumn{2}{c}{\textbf{WOS}} \\ \cmidrule {3-4} 
		& & Micro-F1 & Macro-F1  \\ \midrule
		
		\multicolumn{1}{c}{\multirow{4}{*}{1}}
            & \multicolumn{1}{|l}{HierICRF-T5} &\textbf{62.39}  &\textbf{51.13}  \\
		& \multicolumn{1}{|l}{HierICRF-BERT} &{62.10}  &{49.51}  \\
            & \multicolumn{1}{|l}{HierVerb} &{61.29}  &{47.70}  \\
		& \multicolumn{1}{|l}{HPT} &49.75  &19.78  \\
		& \multicolumn{1}{|l}{HGCLR} &20.10  &0.50   \\
		& \multicolumn{1}{|l}{BERT (Vanilla FT)} &10.78  &0.25   \\
		\midrule
		
		\multicolumn{1}{c}{\multirow{4}{*}{2}} 
            & \multicolumn{1}{|l}{HierICRF-T5} &\textbf{68.40}  &\textbf{60.05}  \\
		& \multicolumn{1}{|l}{HierICRF-BERT} &{68.14}  &{58.81}  \\
            & \multicolumn{1}{|l}{HierVerb} &{67.92}  &{56.92}  \\
		& \multicolumn{1}{|l}{HPT} &60.09  &35.44  \\
		& \multicolumn{1}{|l}{HGCLR} &44.92  &3.23   \\
		& \multicolumn{1}{|l}{BERT (Vanilla FT)} &20.50  &0.34   \\
		\midrule
		
		\multicolumn{1}{c}{\multirow{4}{*}{4}} 
            & \multicolumn{1}{|l}{HierICRF-T5} &{74.94}  &\textbf{68.27}  \\
		& \multicolumn{1}{|l}{HierICRF-BERT} &\textbf{75.22}  &{67.81}  \\
            & \multicolumn{1}{|l}{HierVerb} &{73.88}  &{64.80}  \\
		& \multicolumn{1}{|l}{HPT} &69.47  &53.22  \\
		& \multicolumn{1}{|l}{HGCLR} &68.12  &52.92  \\
            & \multicolumn{1}{|l}{BERT (Vanilla FT)} &67.44  &51.66   \\
		
		\bottomrule[1pt]
	\end{tabular}
\caption
	{
             We further conduct experiments with the \texttt{T5-large} and \texttt{BERT-large-uncased} on WOS. 
	}
\label{tab: model_scale}
\end{table}

\section{Conclusions}
In this paper, we study the challenge of domain-hierarchy adaptation.
We propose a novel framework named HierICRF which elegantly leverages the prior knowledge of PLMs from the perspective of the path routing performed at the language modeling process for better few-shot domain-hierarchy adaptation and can be flexibly applied in any transformer-based architecture.
We perform few-shot settings on HTC tasks and extensive experiments show that our method achieves state-of-the-art performances on 2 popular HTC datasets while guaranteeing excellent consistency performance.
Moreover, our method provides a perspective for hierarchy-based tasks to integrate into a unified instruction tuning paradigm for pre-training.
For future work, we decide to extend HierICRF for effective non-tuning algorithms of LLM.
\appendix

\section*{Acknowledgments}
We thank the reviewers for their insightful comments. This work was supported by National Science Foundation of China (Grant Nos.62376057) and the Start-up Research Fund of Southeast University (RF1028623234).All opinions are of the authors and do not reflect the view of sponsors.

\bibliographystyle{named}
\bibliography{main}

\end{document}